# An empirical evaluation of attention-based multihead deep learning models for improved remaining useful life prediction

Abiodun Ayodeji[a], Wenhai Wang[a], Jianzhong Su[b], Jianquan Yuan[c], Xinggao Liu [a, *]

[a]*State Key Laboratory of Industrial Control Technology, Control Department, Zhejiang University, Hangzhou 310027, P. R. China*
[b]*Tianjin Jinhang Technical Physics Institute, Tianjin 300192, P. R. China*
[c]*Science and Technology on Complex System Control and Intelligent Agent Cooperative Laboratory, Beijing 100074, P. R. China*

**Abstract**
A single unit (head) is the conventional input feature extractor in deep learning architectures trained on multivariate time series signals. The importance of the fixed-dimensional vector representation generated by the single-head network has been demonstrated for industrial machinery condition monitoring and predictive maintenance. However, processing heterogeneous sensor signals with a single-head may result in a model that cannot explicitly account for the diversity in time-varying multivariate inputs. This work extends the conventional single-head deep learning models to a more robust form by developing context-specific heads to independently capture the inherent pattern in each sensor reading. Using the turbofan aircraft engine benchmark dataset (CMAPSS), an extensive experiment is performed to verify the effectiveness and benefits of multi-head multilayer perceptron, recurrent networks, convolution network, the transformer-style stand-alone attention network, and their variants for remaining useful life estimation. Moreover, the effect of different attention mechanisms on the multi-head models is also evaluated. In addition, each architecture's relative advantage and computational overhead are analyzed. Results show that utilizing the attention layer is task-sensitive and model dependent, as it does not provide consistent improvement across the models investigated. The best model is further compared with five state-of-the-art models, and the comparison shows that a relatively simple multi-head architecture performs better than the state-of-the-art models. The results presented in this study demonstrate the importance of multi-head models and attention mechanisms to improved understanding of the remaining useful life of industrial assets.

Keyword: CMAPSS; deep learning, remaining useful life, attention mechanism, predictive maintenance.

* Corresponding author. E-mail address: lxg@zju.edu.cn (X. Liu).

1. **Introduction**

The fourth industrial revolution is data-driven. This is evident in the volume of novel applications of big data from smart manufacturing, smart grid, autonomous self-driven vehicle, and industrial predictive controllers. Access to big data, ensured by recent improvements in advanced sensors, has increased business productivity and efficiency. Big data availability has also motivated interests in the data-driven approach to solving problems in complex industrial systems. One of the problems currently being solve with big data is complex system condition monitoring and predictive maintenance. Predictive maintenance (PdM) is the state-of-the-art maintenance strategy utilized for critical systems in heavy industries such as chemical plants, nuclear power plants, automotive manufacturing and aerospace industry, to reduce downtime, maintenance cost, and ensure critical components' reliability maintainability. One of the most impactful predictive maintenance tasks is the remaining useful life (RUL) estimation of components and systems using degradation information.

The RUL defines the remaining service time left in a component, a critical step to minimize catastrophic failure. The presence of robust multivariate time series signals derived from parallel measurement of hundreds of process variables with diverse sensors has aided the application of many machine learning models for RUL prediction (Berghout, Mouss et al. 2020, Lyu, Ying et al. 2020, Xiang, Qin et al. 2020). Many statistical and data-driven algorithms have been proposed to estimate the RUL of various industrial components. This is informed by previous successes recorded by data-driven and evolutionary algorithms in tasks such as critical system monitoring (Abraham, Grosan et al. 2005, Ayodeji, Liu et al. 2020, Lv, Wang et al. 2020), pattern recognition (Jiang, Xu et al. 2019, Liu, Gu et al. 2019, Liu, He et al. 2020, Wan, Zeng et al. 2020, Zhao, Zeng et al. 2020), object detection and fault forecasting (Liu, Wang et al. 2012, Ayodeji and Liu 2018, Ayodeji and Liu 2019, Dey, Rana et al. 2019, Djeziri, Benmoussa et al. 2019), process and structural health monitoring (He, Xiao et al. 2017, Ayodeji and Liu 2018, Gao and Liu 2018, Ayodeji, Liu et al. 2019, Feng, Borghesani et al. 2019, Feng, Smith et al. 2021), image segmentation (Feng and Chou 2011, Feng, Wong et al. 2018, Agrawal, Panda et al. 2019) and predictive control (Cheng and Liu 2015, Wang, Chen et al. 2017, Boukadida, Benamor et al. 2019, Das Sharma, Chatterjee et al. 2021).

The capability of machine learning models to learn the complexity in noisy, non-linear dataset is being used to provide better insights into the current and future states of processes, components, and systems in heavy industries. Deep neural networks have increasingly been used for multivariate RUL prediction, demonstrating substantial performance improvements over the traditional machine learning models. While many architectures have focused on variants of recurrent neural network (RNN) architectures, recent improvements have also used convolution neural networks, and attention-based models to enhance predictions.

The deep learning approach to prognosis and health monitoring comes with many promises. However, there are also critical weaknesses, especially in its application to learn patterns in multivariate time series. This is because of the dynamic, non-stationary, and spatio-temporal nature of time-series signals. Moreover, in most applications of deep learning models, a network with a single input unit (single-head) is routinely used to extract the features in all the signals in the multivariate dataset. Despite its simplicity, the single-head method assumes that a single unit is robust enough to process all the time series variables effectively. This architecture typically relies on sequential models to encode past inputs, and generate future predictions. However, in many practical scenarios, processing information from heterogeneous sensor networks with a single unit results in a weak model that cannot explicitly account for the diversity in time-varying inputs.

In recent years, two exciting architectures – the multi-head and the attention mechanism – are being used to improve the predictive performance of deep learning models. Unlike other approaches, the multi-head approach utilizes independent "heads" to process each sensor data. This has the advantage of enhanced feature extraction, adjustability ("heads" can easily be added, modified, or removed), and flexibility to new sensor configuration (Canizo, Triguero et al. 2019). The attention mechanism assists the data-driven model to focus more on the informative data segments, and ignore the features that contribute less to the final output. In this way, the underlying relationships between observations and the target can be better explored. Also, the attention mechanism can model the dependencies between the target and the input sequences, and has been successfully applied in many tasks with impressive performance. However, no research has been done to deeply evaluate the effect of these mechanisms jointly or independently on deep learning model, especially for remaining useful life predictions.

The primary goal of this paper is to explore an alternative and optimized way of using deep learning models for RUL prediction. Here, the aim is to combine the knowledge acquired by multiple agents instead of the single agent in the traditional case. To obtain a better result, the multi-head architecture is also embedded with self-attention mechanism. Since there is no consensus on the best network for multivariate prediction, the work investigates the performance of the multilayer perceptron (referred to as the fully connected network in this work), recurrent networks, (including simple recurrent network (SRNN) gated recurrent unit (GRU), long-short term memory (LSTM), and bidirectional LSTM), convolution network, the transformer-style stand-alone attention network, and their ensemble. To properly investigate the model performance, this work answers three crucial questions:

1. Are multi-head models better than single head models for multivariate time series prediction?
2. What effect does the attention mechanism have on multi-head models?

3. Which multi-head model architecture is appropriate to capture the inherent patterns in multivariate time series, such as the turbofan engine run-to-failure signals in the CMAPSS dataset?
4. What level of complexity is appropriate for RUL predictive model with multi-head input networks?

To answer these questions, different experiments are designed to extensively evaluate various deep learning models implemented using the single head, multi-head, and attention-based techniques. The experiments involved training different architectures and comparing them with state of the art. First, different multi-head deep learning networks with suitable inductive biases are designed for each sensor in a multivariate time series. In contrast to the previous RUL prediction approach, different heads are developed to process each sensor reading in the multivariate dataset independently, and then the independent features extracted from each signal are concatenated to predict the component RUL. In addition, an attention mechanism is integrated to extract the context-specific information within each signal. This results in a robust RUL predictive model that gives a state-of-the-art performance on two different subsets of the CMAPSS benchmarked dataset. The current work adds to knowledge by:

1. Presenting the empirical results of the performance evaluation of different multi-head models on multivariate time series signals.
2. Analyzing the effect of self-attention as an ensemble and as a stand-alone model for long sequence time-series prediction tasks.
3. Presenting the exploratory data analysis of the signals in the CMAPSS turbofan dataset.
4. Extensively comparing the multi-head model with single head models to bridge the knowledge gap on their predictive performance for time series signals.

The empirical evaluation results show the importance of multi-head models for critical system safety service life estimation. This work further illustrates the applicability, benefit, and limitations of multi-head models and attention mechanism for predictive maintenance tasks.

## 2. Preliminary

### 2.1. Conventional deep learning models

This section introduces the fundamental deep learning models and different architectures evaluated in this work. The section describes the theoretical background and definitions of the fully connected neural network, recurrent neural networks, convolution network, attention network, and variants commonly used for time series prediction.

#### 2.1.1. Fully connected units (Multilayer perceptron)

A fully connected neural network (FNN) is a simple backpropagating network where each neuron receives input from all the previous layers. The FNN is structurally agnostic universal approximators capable of learning any function. For a fully connected layer with input $x \in R$, the i-th output $y_i \in R$ from the layer is given by:

$$y_i = \sigma(w_i + \cdots + w_n) \tag{1}$$

Where σ is an activation function, and $w_i's$ are learnable parameters in the network. The FNN is defined as the Dense layer in Keras, a flexible API that allows user-defined architecture. The sequential API is utilized to develop the attention-based multi-head FNN architecture as described in section 2.2. Despite its broad applicability, the FNN may have weaker performance than specialized models in some applications. To develop complex models tuned to the structure of the problem, many architectures have been proposed. Other architectures and their implementation in this work are discussed below.

### 2.1.2. Simple Recurrent Neural Networks (SRNN)

Unlike the fully connected units, the simple recurrent neural network performs the same function to each input $x_{(t)}$, and the output $y_{(t)}$ depends on both the input $x_{(t)}$ and the previous hidden state $h_{(t-i)}$. The operation in a simple recurrent unit can be expressed as:

$$h_{(t)} = \sigma(W_{(x)}x_{(t)} + W_{(h)}h_{(t-i)} + b_{(h)}) \qquad 2$$

$$y_{(t)} = \sigma(W_{(y)}h_{(t)} + b_{(y)}) \qquad 3$$

Where $x_{(t)}$ is the input vector, $W$ and $b$ are the learned parameters, $h_{(t)}$ is the hidden state vector, σ is the activation function, and $y_{(t)}$ is the output vector. The recurrent neural network has different architectures commonly used for RUL prediction. Four of those architectures examined in this work are the simple recurrent neural network (SRNN) defined above, the gated recurrent unit, the long-short term memory, and the bi-directional long-short term memory.

### 2.1.2.1. Long-short term memory network

The long-short term memory network (LSTM) is a type of recurrent neural network proposed to overcome the vanishing and exploding gradients common to conventional recurrent neural networks. LSTM can learn and retain order and temporal patterns in long sequences. An LSTM comprises series of information-processing gates controlled by the current values of the input $x_t$ and cell $c_t$ at time $t$, plus some gate-specific parameters. A typical LSTM cell contains the forget gate, the input gate, the output gate, the hidden state, and a memory state. For a given piece of information stored in the network cell, the LSTM works by allowing the input $x_t$ at time t to influence the storing or overwriting of the memory. The input and forget gates decide to keep a new memory or overwrite the old memory. A final output gate determines when to output the value stored in the memory cell to the hidden layer.

For a given input vector $x_{(t)}$, the mathematical formulation of LSTM units comprising the input gate $x_{(t)}$ the forget gate $f_{(t)}$, the output gate $o_{(t)}$, a new memory cell $\bar{c}_{(t)}$, the final memory cell $c_{(t)}$, and the current cell output $h_{(t)}$ is expressed as:

$$i_{(t)} = \sigma(W_{(i)}x_{(t)} + U_{(i)}h_{(t-i)}) \qquad 4$$

$$f_{(t)} = \sigma(W_{(f)}x_{(t)} + U_{(f)}h_{(t-i)}) \qquad 5$$

$$o_{(t)} = \sigma(W_{(o)}x_{(t)} + U_{(o)}h_{(t-i)}) \qquad 6$$

$$\bar{c}_{(t)} = tanh(W_{(c)}x_{(t)} + U_{(c)}h_{(t-i)}) \qquad 7$$

$$c_{(t)} = f_{(t)} * \bar{c}_{(t-1)} + i_{(t)} * \bar{c}_{(t)} \qquad 8$$

$$h_{(t)} = o_{(t)} * tanh(c_{(t)}) \qquad 9$$

Where $h_{(t-1)}$ is the previous cell output, $\bar{c}_{(t-1)}$ is the previous cell memory, and $W$, $U$ are the weight vectors. The capability of LSTM to retain the long- and short-term memory in the cell state and prevent vanishing gradient has been explored in many applications involving time series prediction.

### 2.1.2.2. Gated recurrent unit

The gated recurrent unit (GRU) is a type of recurrent neural network also developed to solve the vanishing gradient problem of the standard RNN. Similar to the LSTM, GRU also uses the gating mechanism to control the memorization process. The GRU has two main gates, a reset gate and an update gate, and uses the hidden state to transfer information. Similar to the forget gate and input gate of LSTM, the GRU's update gate decides if the cell state should be updated with the candidate state (current activation value) or not. The reset gate is used to determine whether the previous cell state is essential or not. The reset gate stores the relevant information from the past time step into the new memory content. Then it multiplies the input vector and hidden state with their weights. After that, the units perform element-wise multiplication between the current reset gate and previously hidden state. The result is summed, and a non-linear activation function is applied to produce an output. The candidate cell is similar to the hidden state(activation) of a conventional recurrent unit, and the update gate transfers information from the current unit to the rest of the network. In GRU, the final cell state is directly passing as the activation to the next cell, and its architecture is less complex and computes faster than LSTM.

### 2.1.2.3. Bidirectional LSTM

The bidirectional LSTM is an extension of the traditional LSTM that can improve model performance on sequential problems. This structure allows the networks to have both backward and forward information about the sequence at every time step. When all time steps of the input sequence are available, a BiLSTM trains two instead of one LSTM on the input sequence. This provides additional context to the network and results in faster learning.

One disadvantage of traditional LSTM lies in that only preceding information is utilized for computing. BiLSTM can address the problem by using two separate hidden layers: the hidden forward sequence $\overrightarrow{h_t}$, and the backward hidden sequence $\overleftarrow{h_t}$. The BiLSTM combines the $\overrightarrow{h_t}$ and $\overleftarrow{h_t}$ to generate the output $y_t$. Given a hidden state h of an LSTM block, the BiLSTM is implemented with the following function:

$$\overrightarrow{h_t} = H(W_{x\overrightarrow{h}} x_t + W_{\overrightarrow{h}\overrightarrow{h}} \overrightarrow{h}_{t-1} + b_{\overrightarrow{h}}) \qquad 10$$

$$\overleftarrow{h_t} = H(W_{x\overleftarrow{h}} x_t + W_{\overleftarrow{h}\overleftarrow{h}} \overleftarrow{h}_{t-1} + b_{\overleftarrow{h}}) \qquad 11$$

$$y_t = W_{\overrightarrow{h}y} \overrightarrow{h}_t + W_{\overleftarrow{h}y} \overleftarrow{h}_t + b_y \qquad 12$$

Where all the notations retain their previous definition.

### 2.1.3. One-dimensional convolution network

Some of the most widely-used deep learning models are developed with the convolution neural network (CNN). CNNs are used to capture spatially invariant features in images and patterns in text. Many CNN-enabled deep learning models have also been proposed to capture patterns and structures in the time series dataset used for RUL prediction tasks. For a one-dimensional convolution neuron(1D-CNN) in layer $l$, with a previous layer and next layer defines as $l-1$ and $l+1$ respectively, the input of the $k^{th}$ neuron in layer $l$ can be expressed as (Kiranyaz, Gastli et al. 2018):

$$x_k^l = b_k^l + \sum_{i=1}^{N_{l-1}} conv1D(w_{ik}^{l-1}, s_i^{l-1}) \qquad 13$$

where $w_{ik}^{l-1}$ is the weight of the 1D kernel from the $i^{th}$ neuron at layer $l-1$ to the $k^{th}$ neuron at layer $l$, $x_k^l$ is the input, $b_k^l$ is the bias term, and $s_i^{l-1}$ is the output of the $i^{th}$ neuron at layer $l-1$. For a network with input layer $l$, input vector $p$, output layer $L$, and the corresponding output vector $[y_1^L, ..., y_{N_L}^L]$ the objective is to minimize the error between the input and the output for every epoch of the input signal. Details of one-dimensional convolution network characteristics and inner properties can be found in previous discussions (Kiranyaz, Gastli et al. 2018).

A common approach to obtain an optimal model is integrating the model outputs. Some implementation stacks conventional CNN with recurrent layers to improve model accuracy. In this work, a different architecture composed of the hybrid of CNN and LSTM is also explored. The new architecture, referred to as the temporary convolution network (CNLSTM), utilizes memory units and temporal pattern capturing capability of LSTM with the spatially invariant feature capturing capability of CNN to predict the RUL.

### 2.2. Multi-head architecture and attention mechanism

The networks discussed in Section 2.1 above have been used for different tasks. However, their multi-head implementation and the effect of attention mechanism on the multi-head architecture have not been fully explored for RUL prediction. This section discusses the multi-head implementation and the attention mechanism used in the experimental evaluation of the deep learning models. This section also describes different salient neural architectures evaluated.

#### 2.2.1. The multi-head mechanism

In multivariate time series prediction, some variables are independent of each other and thus not correlated. This uncorrelated information is expected in heterogeneous sensor systems that capture different process variable at different time scales or frequencies. Hence, it is crucial to develop a specialized model to process this information separately and capture the local and global features inherent in the data. The multi-head architecture combines multiple network structures, in which each head learns features in different representation subspaces, leading to multi-representation that further improves the predictive model performance. Each sensor output is processed on a fully independent head in a multi-head network, responsible for extracting meaningful features from the sensor data. In addition, this enables the spatial representation capturing, and the output in each subspace is concatenated to enhance the contextual information in each time series.

Besides, processing each sensor data on an independent head has other advantages (Canizo, Triguero et al. 2019):

(1) The feature extraction is enhanced by focusing only on one particular sensor rather than all at once.
(2) Each network head can be adjusted to the specific nature of each sensor data, and
(3) It results in a flexible architecture that is adaptable to new sensor configurations.

For a deep learning model to properly fit the multi-head implementation, the input data is preprocessed differently. The input is shaped such that the number of heads in the model matches the number of signals in the input. A similar data preprocessing is done for the test data used for model evaluation. In this work, the conventional models defined in section 2.1 are developed with multiple heads, serving each input sequence in the dataset.

### 2.2.2. Attention mechanism

The attention mechanism is used initially to resolve a critical demerit of fixed-length encoding of context vector design, resulting in compression and loss of information in sequence to sequence recurrent networks. For a long sequence, time series task, the network often forgets the first part once it completes processing the whole input. In natural language processing, the attention vector is used to estimate how strongly a word is correlated with other elements and take the sum of their values weighted by the attention vector as the target approximation. Consider a recurrent encoder-decoder network, where the encoder encodes a sequence of input vector $x = (x_1, ..., x_n)$ into a context vector $\hat{c}$. The hidden state of the recurrent encoder at time $t$ is given by:

$$h_t = f(x_t, h_{t-1}) \qquad 14$$

And the context vector generated from the hidden state sequence $\hat{c}$, is given by:

$$\hat{c} = q(\{h_1, ..., h_n\}) \qquad 15$$

Where $f$ and $q$ are nonlinear functions. Given the context vector, and the previous sequence $\{y_1, ..., y_{t-1}\}$, the decoder predicts the next sequence $y_t$, by decomposing the joint probability, such that:

$$p(y) = \prod_{t=1}^{T} p(y_t | \{y_1, ..., y_{t-1}\}, \hat{c}) \qquad 16$$

Where the decoder output vector $y = (y_1, ..., y_m)$. In the context of attention mechanism, each conditional probability expressed in equation 16 above is defined as:

$$p(y_i | y_1, ..., y_{i-1}, x) = g(y_{i-1}, a_i, \hat{c}_i) \qquad 17$$

Where $g$ is a nonlinear function, and $s_t$ is the attention vector of the hidden state at time $t$, given as:

$$s_t = f(s_{t-1}, y_{t-1}, \hat{c}_t) = f(\hat{c}_t, h_t) \qquad 18$$

Hence, the context vector $\hat{c}_i$ computed as a weighted sum of the sequence of annotations $h_t = (h_1, ..., h_n)$, is given by:

$$\hat{c}_t = \sum_{j=1}^{J} \alpha_{tj} h_j \qquad 19$$

Where $\alpha_{tj}$ is the attention weight from the t-th output to the j-th input, and $h_j$ is the encoder state for the j-th input. The attention weight is expressed in terms of the alignment model (attention score, $e_{tj}$) as:

$$\alpha_{t,j} = \frac{\exp(score(\mathbf{h}_t,\mathbf{h}_j))}{\sum_{j'=1}^{J}\exp(score(\mathbf{h}_t,\mathbf{h}_{j'}))} \qquad 20$$

The alignment model, defined as $e_{tj} = f(\mathbf{s}_{t-1}, \mathbf{h}_j)$ is the measure of the distance between the inputs around position $t$ and the output position, and $f$ is the alignment model which scores the input-output matching distance, and $\mathbf{s}_{t-1}$ is the hidden state from the previous timestep. A detailed description of the attention annotations can be found in (Bahdanau, Cho et al. 2014).

The attention mechanism has different variants, distinguished by the alignment score computation approach. One of the most common variants, self-attention, also has an additional hidden state, where each hidden state attends to the previous hidden states of the same model (Lin, Feng et al. 2017). Moreover, the self-attention network can be implemented as soft attention or hard attention, depending on the alignment score. The soft attention architecture considers the global context for each time stamp (Bahdanau, Cho et al. 2014), and the attention alignment weights are learned and laced over all patches in the input. One advantage of this approach is that it makes the model differentiable. However, it is expensive when the source input is large. For the hard attention network, the weights only select one patch of input to attend to at a time. This implementation has fewer calculations at inference time. However, the model is non-differentiable and may require more complex techniques such as variance reduction to train.

In this work, the attention types are classified in term of how the alignment score $(h_t, h_s)$ is computed, as shown in Table 1. For a given number of hidden states $H$ and trainable weight matrices $W_a$ and $v_a$, target hidden state $h_t$ and a source hidden state $h_s$, a different approach to computing the scores $(h_t, h_s)$ have been explored in the literature. These approaches result in differences in how attention is utilized and significantly affect the model's predictive performance. Table 1 shows the common attention mechanism and their alignment score functions.

Table 1. Different attention types and their alignment score function

| Name | Alignment score function | Reference |
|---|---|---|
| Dot product | $h_t^T . h_s$ | (Luong, Pham et al. 2015) |
| Scaled dot product | $h_t^T . h_s / \sqrt{H}$ | (Graves, Wayne et al. 2014) |
| Additive attention | $v_a^T . \tanh(W_a[h_t : h_s])$ | (Bahdanau, Cho et al. 2014) |
| Content-based attention | $cosine[h_t . h_s]$ | (Graves, Wayne et al. 2014) |
| General | $h_t^T . W_a . h_s$ | (Luong, Pham et al. 2015) |
| Location-based | $W_a . h_t$ | (Luong, Pham et al. 2015) |

Since its introduction, the attention mechanism has been widely applied in machine translation, natural language processing, sentiment classification, text generation etc. However, their implementation on long sequence time series prediction is rare. This work utilizes the Keras implementation of the self-attention network. Specifically, the architecture defines a self-attention mechanism that computes the hidden alignment scores ($h_{t,t'}$), the attention weight ($a_t$), the alignment model ($e_{t,t'}$) and the context vector ($l_t$) as:

$$h_{t,t'} = \tanh(x_t^T W_t + x_{t'}^T W_x + b_t) \qquad 21$$

$$e_{t,t'} = \sigma(W_a h_{t,t'} + b_a) \qquad 22$$

$$\alpha_t = softmax(e_t) \qquad 23$$

$$l_t = \sum_{t'} \alpha_{t,t'} x_{t'} \qquad 24$$

where *W*'s and *b*'s are weights and biases to be learned. Moreover, for all self-attention modules, two different attention score computation is provided, defined as:

$$score(\boldsymbol{h}_t, \boldsymbol{h}_j) = \begin{cases} \boldsymbol{h}_t^T \boldsymbol{W} \boldsymbol{h}_j & [Multiplicative\ attention] \\ \boldsymbol{v}_a^T \tanh(\boldsymbol{W}_1 \boldsymbol{h}_t + \boldsymbol{W}_2 \boldsymbol{h}_j) & [Additive\ attention] \end{cases} \qquad 25$$

Where $\boldsymbol{W}$, $\boldsymbol{W}_1$ and $\boldsymbol{W}_2$ are weight matrices and $\boldsymbol{v}_a^T$ is a weight vector. In this study, the multiplicative attention score and the soft and hard attention layers are evaluated in the experiments. In addition, early investigations show that using pure self-attention underperforms, as opposed to regularized attention. That is attributed to the fact that pure self-attention loses rank exponentially with depth. Hence, a regularized self-attention, with the regularizer defined as $||AA^T - 1||_F^2$ is used. Moreover, based on early experiments, a constant value of 1e-4 is selected as the kernel, bias, and attention regularizer for all attention-based experiments investigated in this paper.

This work investigates the effect of the attention mechanism by introducing a different source of information provided by the attention module on top of the multi-head implementation of MLP (FNN), SRNN, CNN, LSTM, BiLSTM, CNLSTM, and a stand-alone attention model (SAN). The stand-alone self-attention utilized in this work is a transformer-styled attention model entirely built on the self-attention network without using the recurrent or convolution layers. To guide the extraction of time series features, and reduce the risk of overfitting, the leaky-relu and BatchNormalization layers are also utilized, forming the innovative architectures shown in Fig. 1-3. Fig 1-3 illustrates the architecture of multi-head models with five input signals. It is worth noting that the CMPASS dataset is a multivariate time series, with a total of twenty-six signals in each data subset, and the architecture used for each subset has the same number of heads as the input signals.

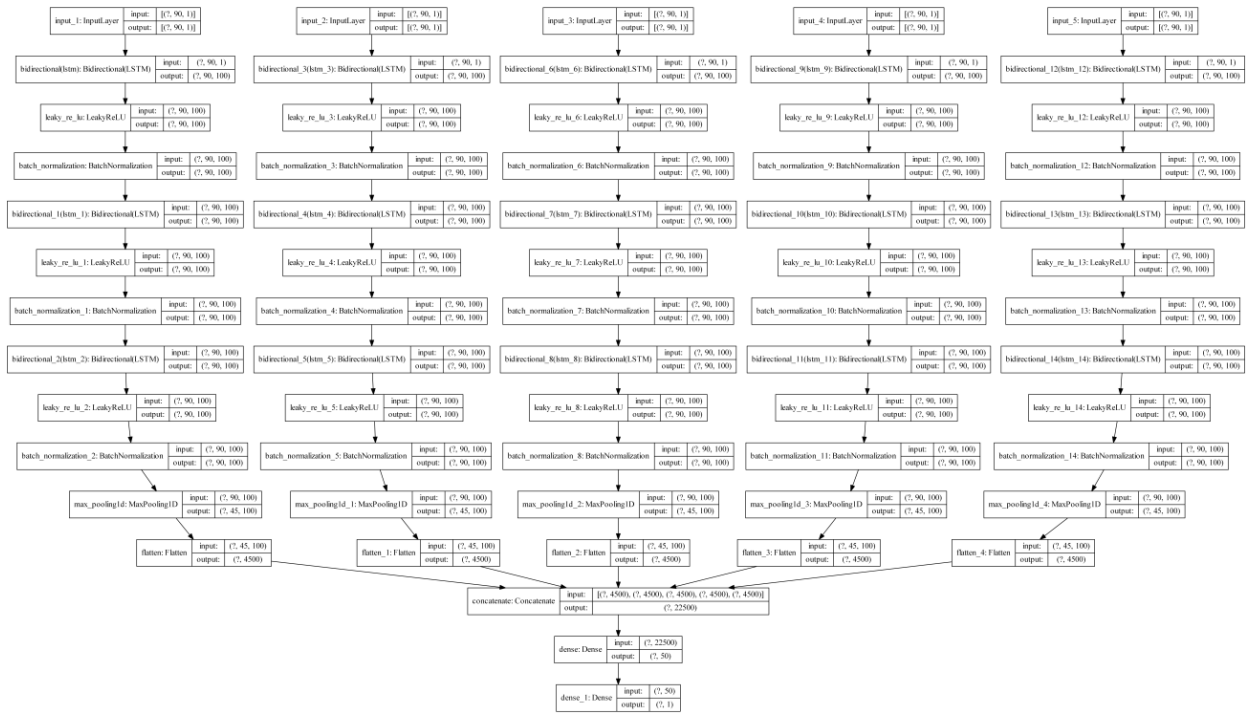

Fig 1: Attention-based bidirectional LSTM architecture

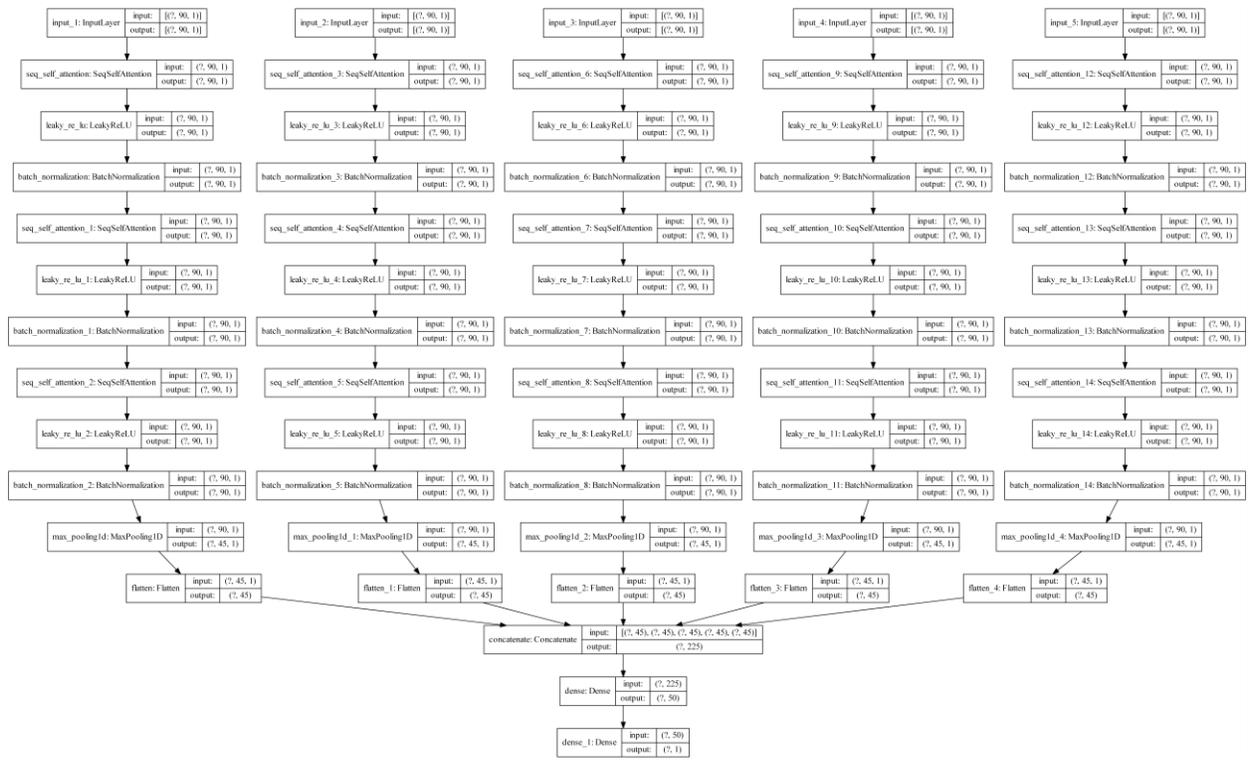

Fig 2: Stand-alone self-attention architecture.

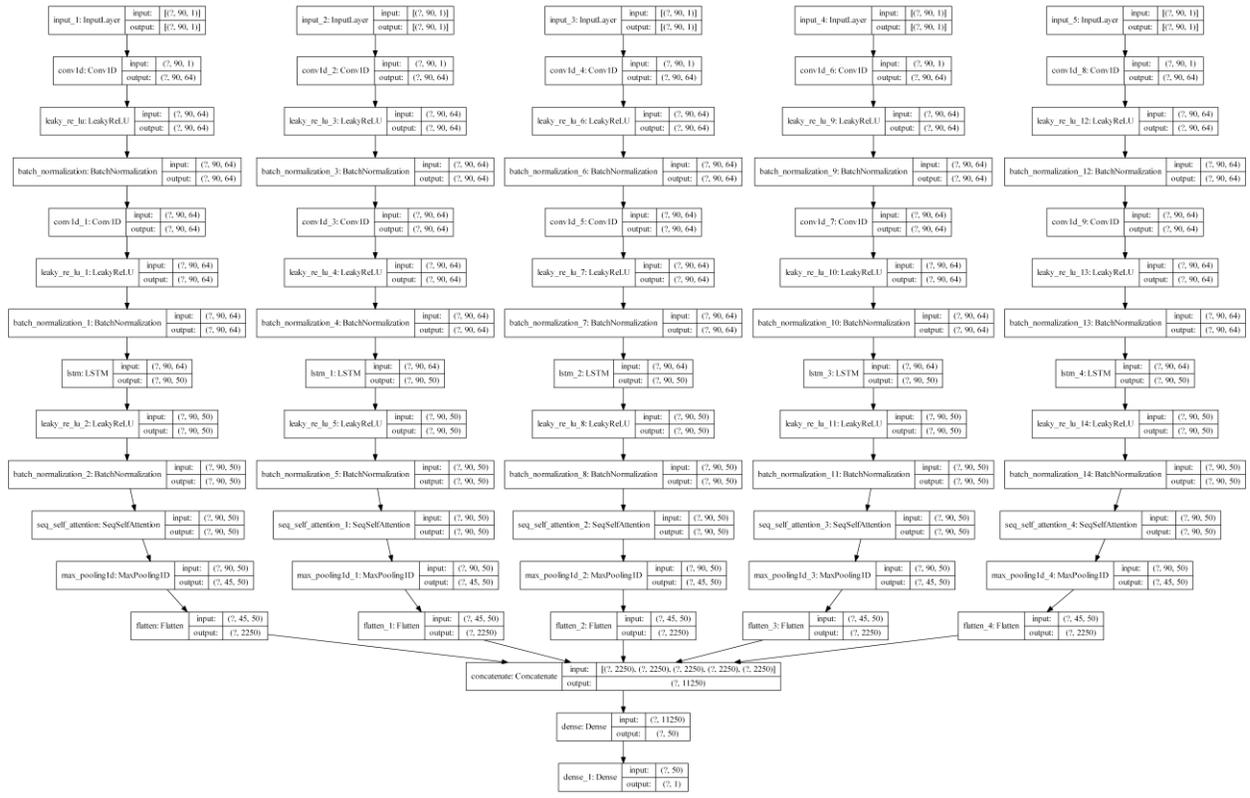

Fig 3: Attention-based multi-head convolution architecture

## 3. Data description and preprocessing

### 3.1. Exploratory data analysis and sensor selection

The self-attention-based multi-head models are evaluated on the FD001 and FD003 in the Commercial Modular Aero-Propulsion System Simulation (C-MAPSS) dataset that defines a turbofan aircraft engine degradation prognostic benchmarking problem. The dataset comprises a multi-variate time series with 26 variables, including the 21 sensor measurements, cycles, sensor settings, and engine identification numbers. The composition in CMAPSS is a record of run-to-failure degradation patterns in multiple airplane engines collected under different operational conditions and fault modes. Each engine unit starts with varying degrees of initial wear, and variations in the operation of the engine units introduced wear which is subsequently captured in the dataset. The distribution of some of the original signals in the FD001 and FD003 datasets is shown in Fig. 4-7, and a comprehensive description of the dataset composition, acquisition path, and the turbofan engine flight condition indicators can be found in (Saxena, Goebel et al. 2008).

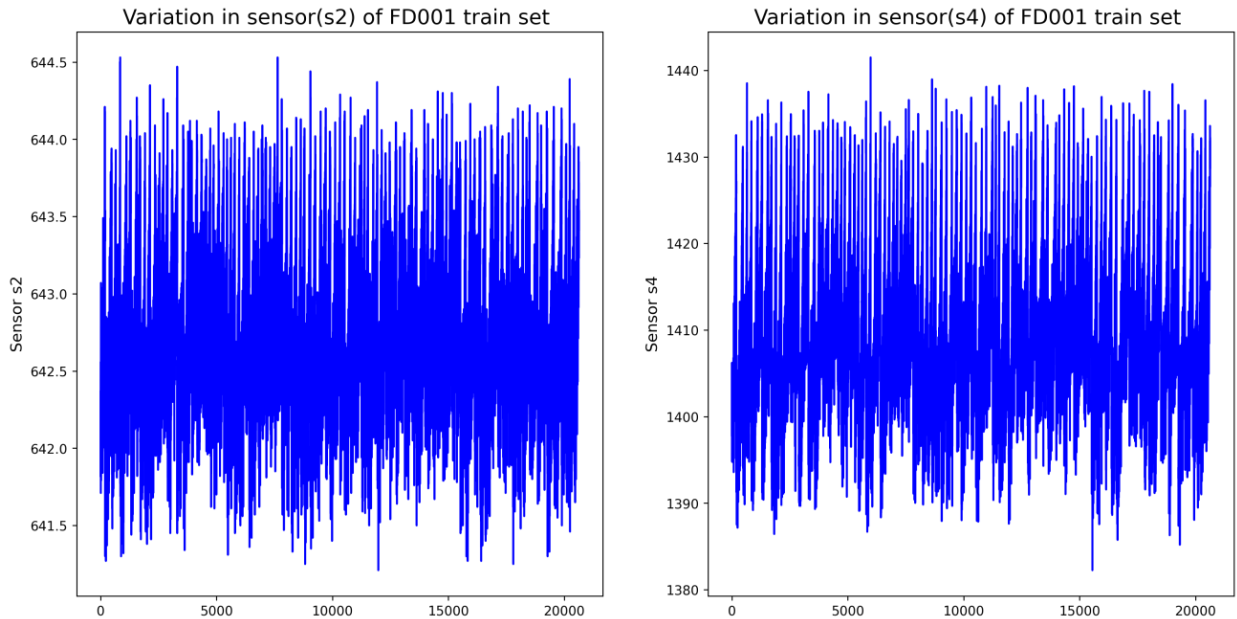

Fig 4: The waveform of absolute values of sensor signal s2 and s4, in FD001

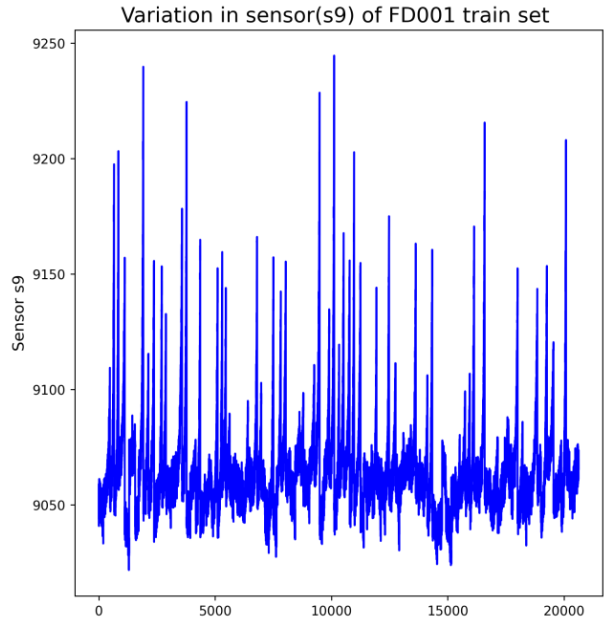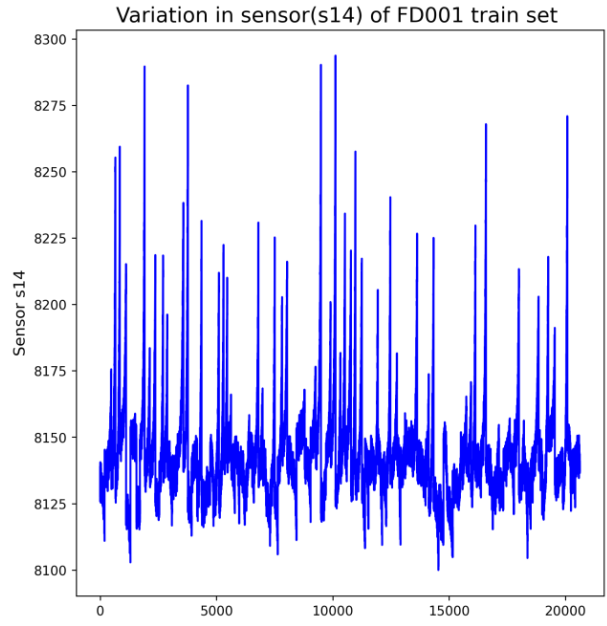

Fig 5: The waveform of absolute values of sensor signal s9 and s14 in FD001

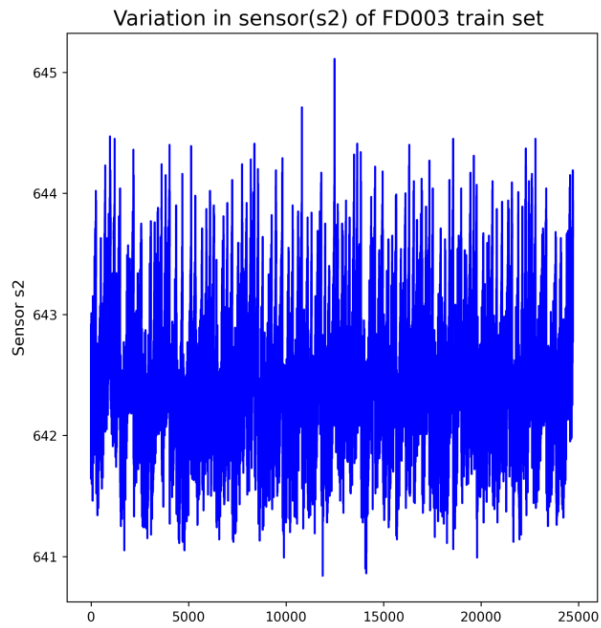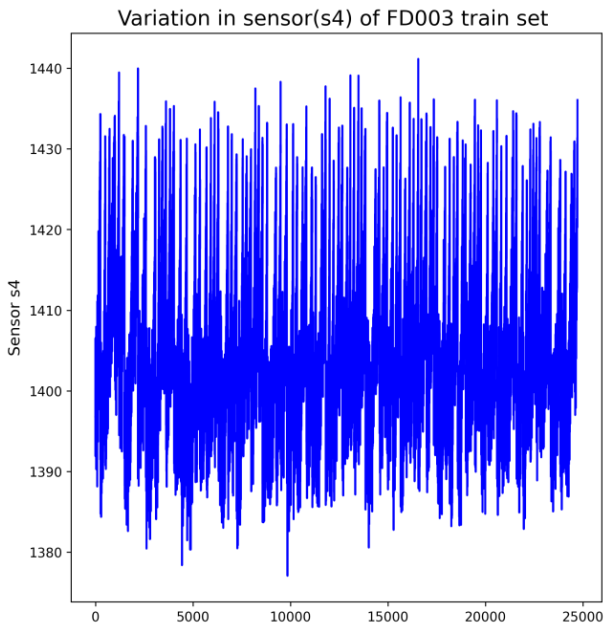

Fig 6: The waveform of absolute values of sensor signal s2 and s4 in FD003

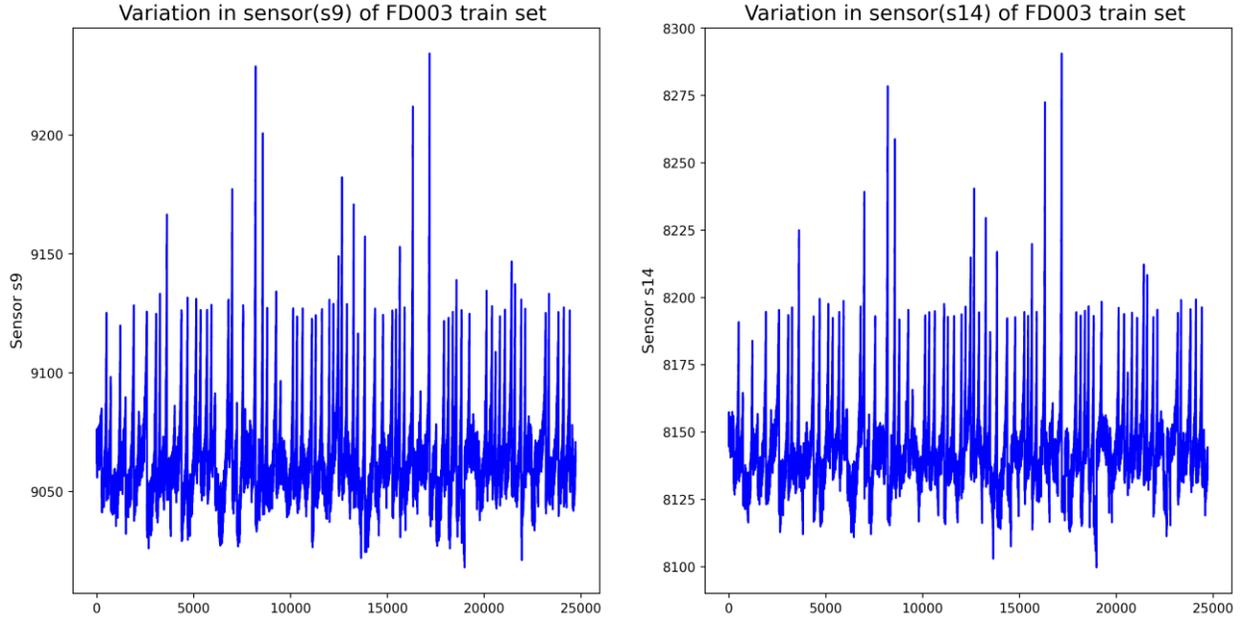

Fig 7: The waveform of absolute values of sensor signal s9 and s14 in FD003

The inherent information in the CMAPSS dataset has been used to predict the turbofan engine remaining useful life. Previous works have shown that some sensor measurements do not provide additional information to aid the prognostic task (Chen, Jing et al. 2019, Ellefsen, Bjørlykhaug et al. 2019, Wen, Zhao et al.). However, the previous works did not explain the reasoning that informed the sensors discarded. Hence, to improve the understanding of the dataset and aid reproducibility, this section discusses the exploratory data analysis for the subsets in CMAPSS dataset to show the decision behind the sensors selected to have the essential information for model training. Fig. 8 and 9 below show the distribution of each signal in data subset FD001 and FD003, respectively. In FD001, it is observed that sensors [1,5,6,10,16,18,19] and setting 3 have features with no learnable pattern. Selecting these features would result in a complex model that is computationally expensive. Hence, these sensors are discarded from FD001, leaving 17 inputs. Similarly, for dataset FD003, it is observed that the features presented in sensors [1,5,16,18,19] and setting 3 do not have any learnable distribution. The signals do not have useful information that would aid the model's predictive capability and are therefore discarded.

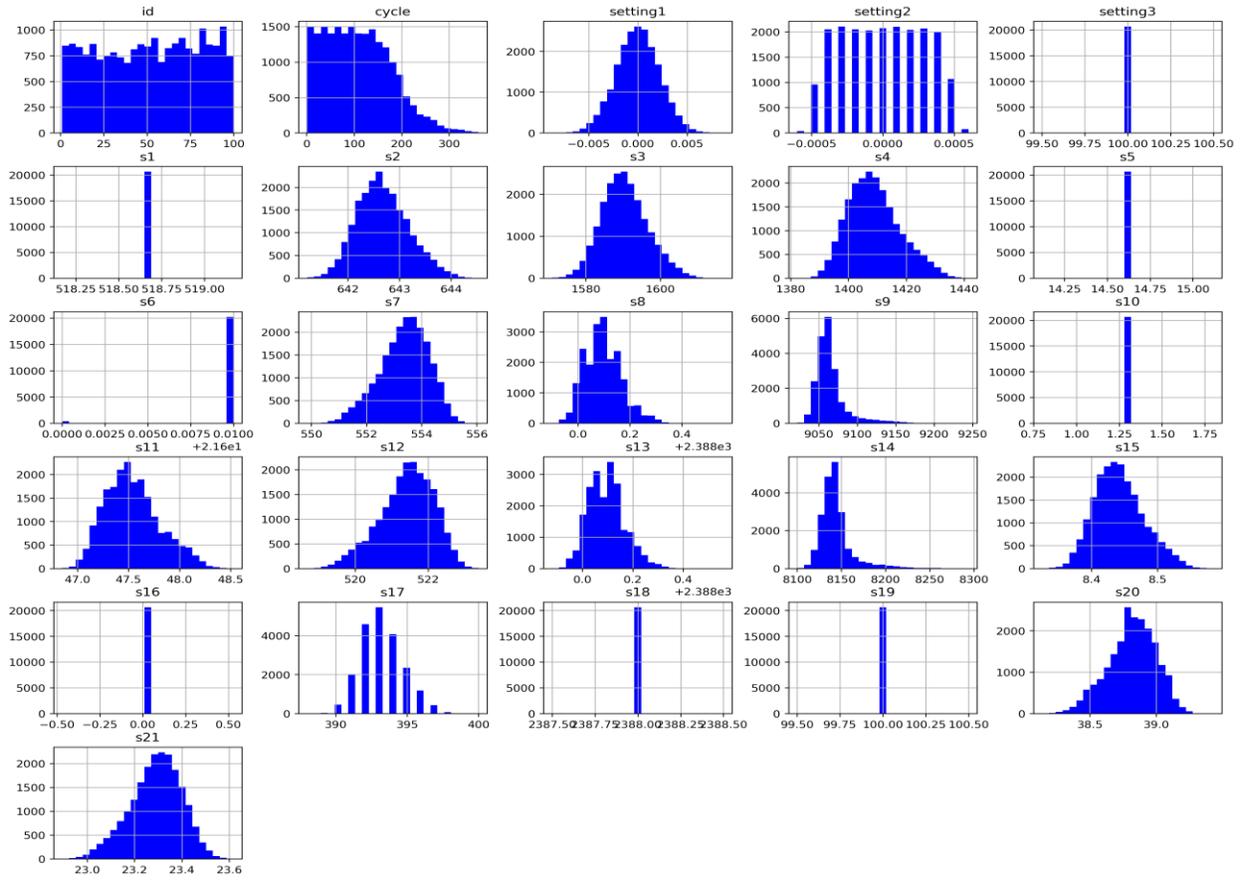

Fig 8: Condition indicators in FD001 dataset

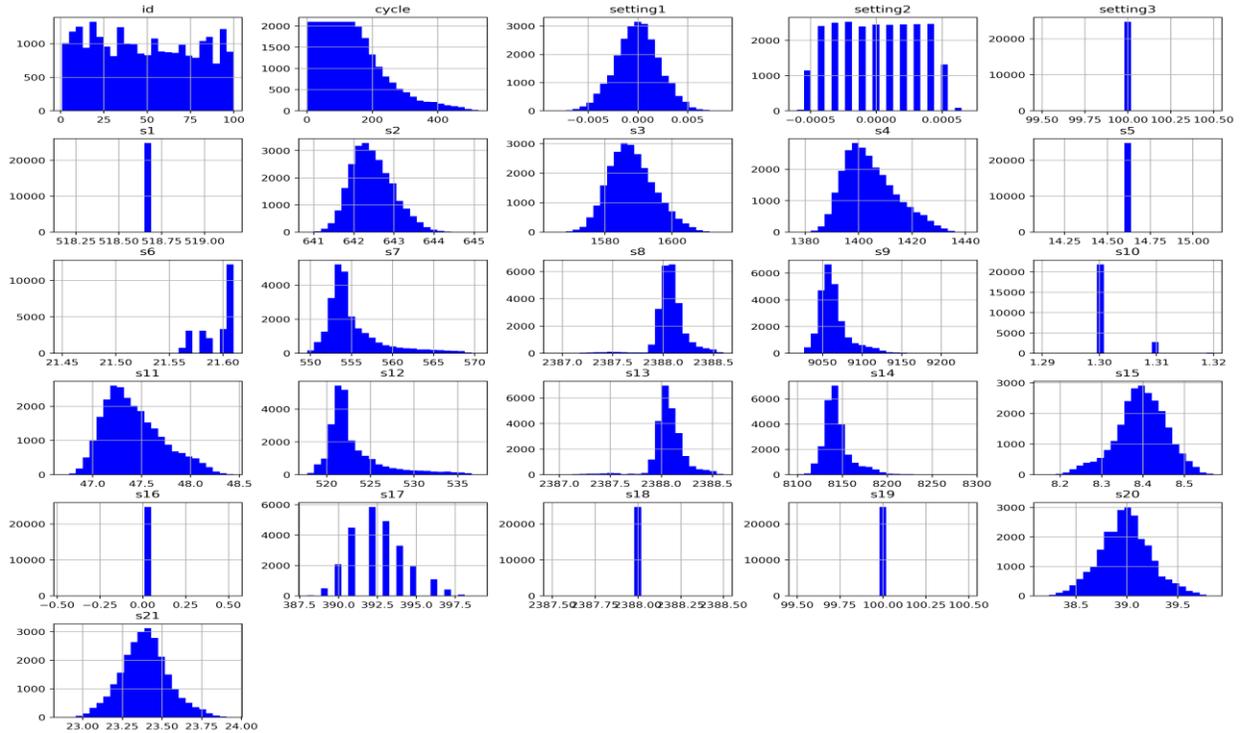

Fig 9: Condition indicators in FD003 dataset

## 3.2. Data augmentation and training technique

Since the model aims to predict the RUL for each engine in the dataset, an artificial signal is constructed with the number of cycles in the dataset. The artificial signal indicates the remaining useful life for each timestep in the dataset, and it is used as the training target. In addition, previous studies show that critical components rarely degrade at the beginning of life, i.e, component degradation begins at a particular point in the operating life. The concept is used to augment the artificial signal that represents the RUL. The concept is called a piece-wise RUL approach, and it enhances the predictive performance of the model. To properly select the optimal piece-wise value for all the engine, the distribution of the number of cycles in each engine needs to be known. Fig. 10 and 11 show the cycles distribution for each engine in data subset FD001 and FD003.

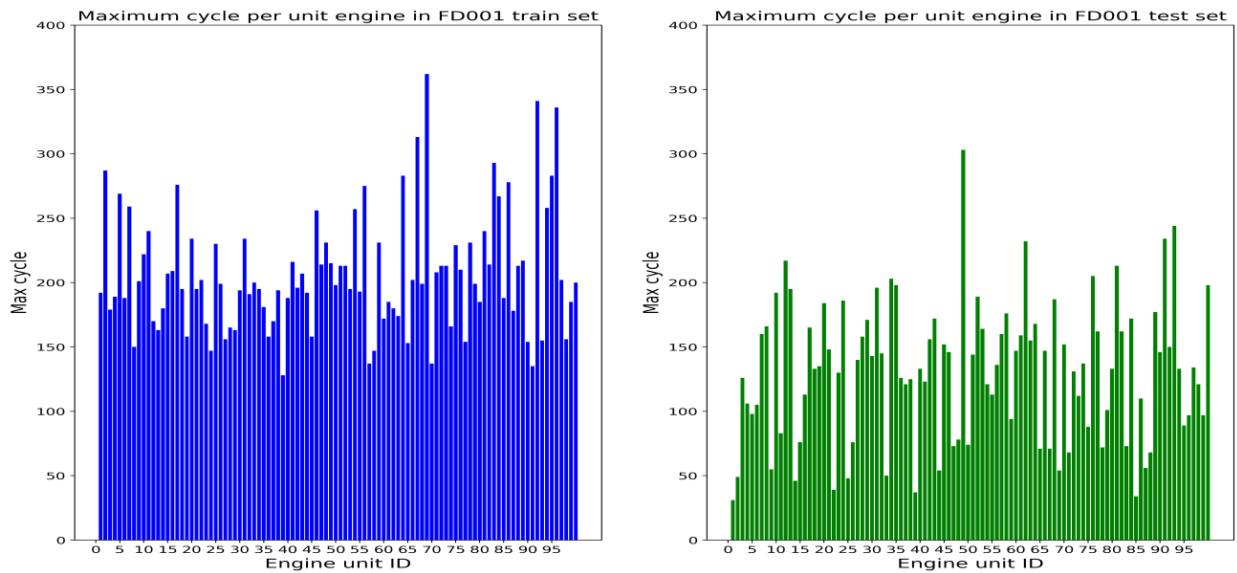

Fig 10: The distribution of engine cycles in FD001 dataset

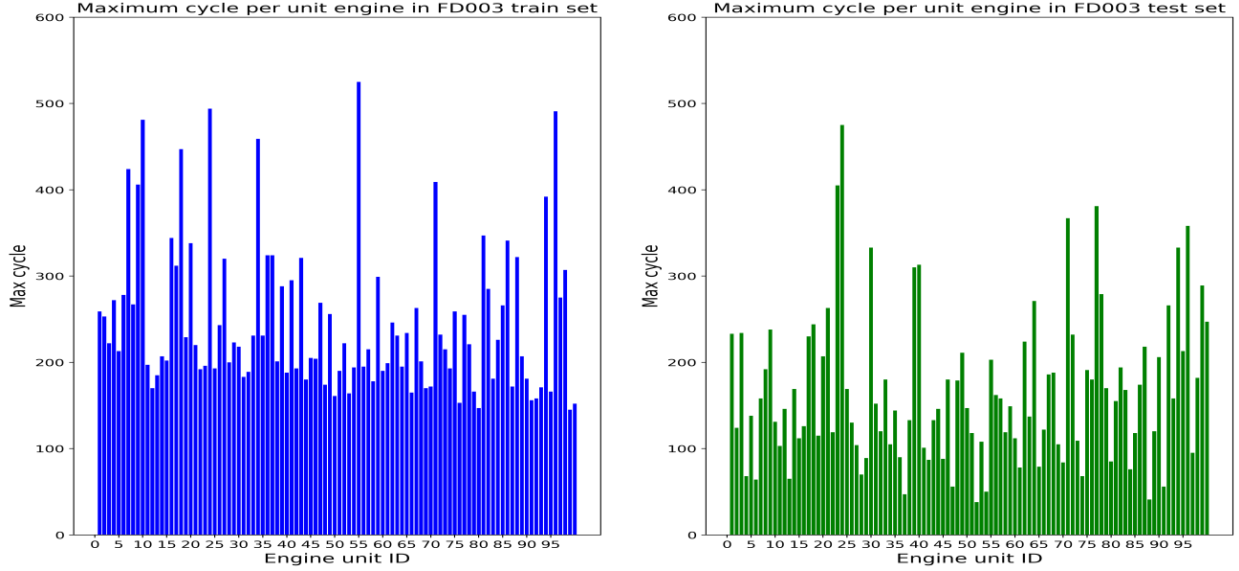

Fig 11: The distribution of engine cycles in FD003 dataset

Fig 10 and 11 confirm that the piece-wise RUL concept is implementable at 130 cycles, as no engine fails under 120 life cycles in both FD001 and FD003 train set. Hence, for each engine, a constant value of 130 ($R_{early}$ =130) is selected as the point at which each engine begins to degrade. To allow the network to learn the optimal parameter from the input node quickly and speed up convergence, the selected signals are normalized and transformed using the Scikitlearn's *MinMax Scaler*, which rescales the input in the range [0, 1]. Then a data generating function is applied to generate the sequences in the form [*Df, Seq_l, Seq_c*], where the *Df is* the preprocessed data frame, *seq_l* is the selected time window (sequence length), and the *seq_c* is the column that represents the desired signals from each data subset.

### 3.3. Metric for model evaluation

There are two commonly used metrics to evaluate models trained on the CMAPSS dataset. These are the root mean squared error (RMSE) and the Score metric, mathematically expressed as:

$$Score = \begin{cases} \sum_{i=1}^{n}(\exp(-e_i/13) - 1), & if\ e_i < 0 \\ \sum_{i=1}^{n}(\exp(-e_i/10) - 1), & if\ e_i \geq 0 \end{cases} \qquad 26$$

$$RMSE = \sqrt{\frac{1}{n}\sum_{i=1}^{n} e_i^2} \qquad 27$$

where $e_i$ is the difference between the estimated RUL and the actual RUL ($RUL_{true} - RUL_{predicted}$) for the *ith* test unit. The training objective is to develop a model that minimizes these metrics, such that late predictions ($e_i>0$) are more heavily penalized than early predictions ($e_i<0$). Moreover, the Score metric penalizes the model with diverging predicted RUL away from the true RUL. This builds a form of cost sensitivity in the model, as false predictions could have catastrophic consequences. The model development method, evaluation routine, and optimization approach are shown in the flow chat in Fig 12.

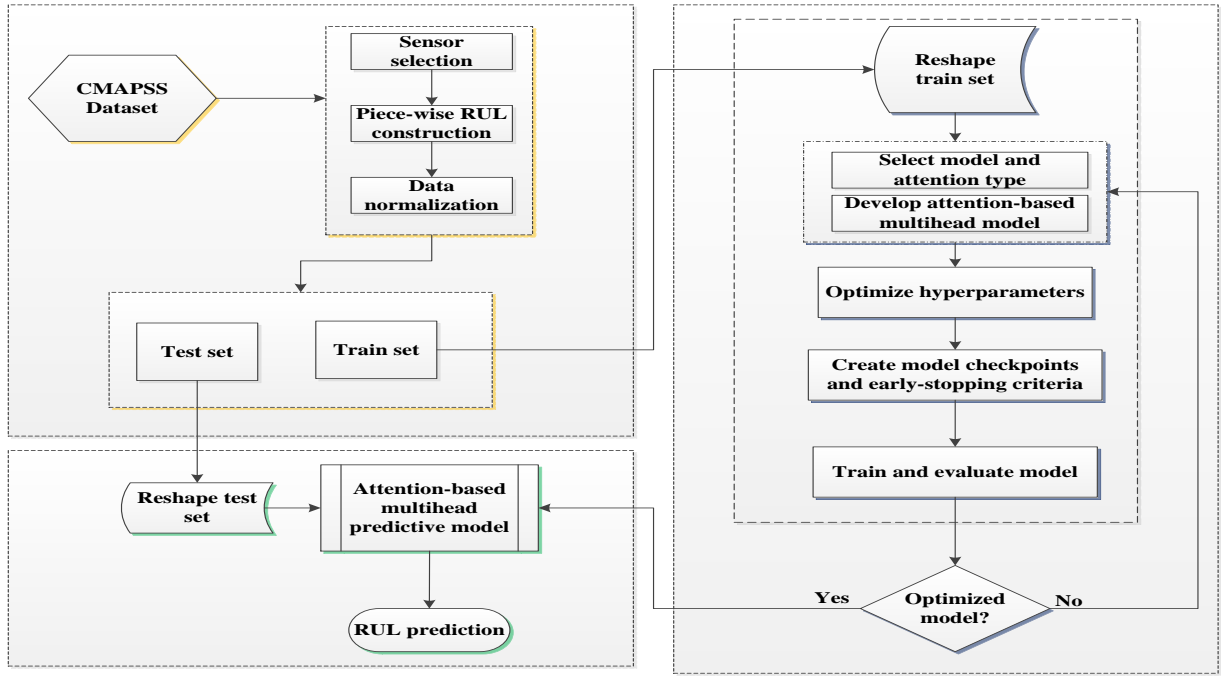

Fig. 12: Self-attention based multi-head model implementation for RUL prediction experiments

## 4. Result and analysis

This section contains the empirical evaluation results of conventional deep learning models and attention-based multi-head deep learning models for RUL prediction. The performance of different configurations and variants of the models is also analyzed, presenting a useful empiricism necessary to further apply multi-head and attention mechanism, especially for multivariate time series prediction tasks. The models evaluated in this section fundamentally contains fully connected neural network (FNN), simple recurrent neural network (SRNN), gated recurrent units (GRU), long-short term memory network (LSTM), convolution neural network (CNN), self-attention network (SAN) and their variants and hybrids. The experiments are performed using TensorFlow with Keras backend, and the models are trained on the intel core i7 workstation running RTX2060s GPU. All experiments are performed with a constant time window length of 90 (i.e., $Seq\_l$, =90).

### 4.1. Single-head vs. multi-head

Tables 2 and 3 presents the evaluation result of eight different single-head and multi-head architectures. This is to demonstrate the performance improvement presented by the multi-head implementation of deep learning models in contrast to the conventional single head. The analysis focuses on the traditional metrics (RMSE and Score values) and the computational burden (number of parameters and GPU time).

In table 2, it is observed that the single-head FNN model's RMSE and score value reduced by 45.65%, and 79.62 %, respectively, between the single-head and multi-head implementation. However, the multi-head model has a significant computational burden. As seen in table 2, the multi-head implementation has 1587.2% more parameters and trains 459.59% slower than the single head model. A similar trend is observed for the rest of the models evaluated with FD001, except the SAN model. For SRNN, multi-head error decreased by 45.37%, and the score improved by 82.42%, compared to the single head model.

Further, the GRU's multi-head error reduced by 38.95%, and the score also reduced by 82.78% compared to the single head. For LSTM, BiLSTM, CNN and CNLSTM, models, the multi-head RMSE values decreased by 37.42%, 0.81%, 53.16%, and 46.45% while the score values reduced by 84.05%, 75.20%, 87.76%, 92.76% respectively. For the SAN exception, the multi-head error increased by 38.2%, and the score also increased by 368.11%. However, all the multi-head models recorded significant computational burdens, as the trainable parameters of the multi-head models increased by 2509.7%, 2054.73%, 1924.98%, 1765.91%, 5974.76%, and 1575.18% for SRNN, GRU, LSTM, BiLSTM, CNLSTM, respectively. However, the SAN multi-head model has a 2.03% reduction in the trainable parameters compared with the single head model. Also, the GPU time consumed by each of the multi-head models is significant in contrast to the single head implementation. It is observed that the SRNN-SAN multi-head models consume 503.17%, 324.04%, 134.39%, 570.72%, 321.07%, 269.14%, and 62.16% more GPU time than their single head implementation.

A similar trend is observed when the models are evaluated in the FD003 dataset, as shown in Table 3. It is seen that the FNN, SRNN, GRU, LSTM, BiLSTM, CNN, and CNLSTM, multi-head models have 38.63%, 19.14%, 20.93%, 26.50% 10.59%, 37.36%, and 27.33% reduction in the RMSE error, and 48.08%, 81.57%, 14.25%, 45.09%, 139.48%, 70.39%, and 70.00% reduction in the score value respectively. In addition, the SAN multi-head model also performed worse than the single-head architecture, with 43.547% increase in the RMSE, and 1273.9% increase in the score value. Moreover, the model's multi-head implementation has 2372.88%, 3719.14%, 3037.21%, 2843.66%, 2618.023%, 6333.62%, and 2346.409% more parameters and spent 147.59%, 616.76%, 935.58%, 422.82%, 1686.07%, 1712.54%, and 590.64% more gpu time respectively.

Table 2: Single-head vs. multi-head deep learning models on FD001 dataset

| Model | Single-head | | | | Multi-head | | | |
|---|---|---|---|---|---|---|---|---|
| | RMSE | Score | parameters | GPU time(s) | RMSE | Score | parameters | GPU time(s) |
| FNN | 15.97 | 139.51 | 119,201 | 51.83 | 8.68 | 28.43 | 2,011,201 | 290.04 |
| SRNN | 18.29 | 200.90 | 81,951 | 122.60 | 9.99 | 35.31 | 2,138,701 | 739.49 |

| Model | | | | | | | | |
|---|---|---|---|---|---|---|---|---|
| GRU | 16.48 | 201.06 | 119,651 | 482.04 | 10.06 | 34.61 | 2,578,151 | 2044.05 |
| LSTM | 15.87 | 187.64 | 137,601 | 546.80 | 9.93 | 29.93 | 2,786,401 | 1281.66 |
| BiLSTM | 14.98 | 157.85 | 335,101 | 368.56 | 14.86 | 39.14 | 6,252,701 | 2472.02 |
| CNN | 18.66 | 217.35 | 45,189 | 92.82 | 8.74 | 26.60 | 2,745,125 | 390.84 |
| CNLSTM | 21.70 | 749.12 | 146,809 | 371.42 | 11.62 | 54.22 | 2,459,321 | 1371.06 |
| SAN | 19.36 | 187.09 | 39,425 | 392.33 | 26.76 | 875.78 | 38,623 | 636.21 |

Table 3: Single-head vs. multi-head deep learning models on FD003 dataset

| | Single-head | | | | Multi-head | | | |
|---|---|---|---|---|---|---|---|---|
| Model | RMSE | Score | parameters | GPU time | RMSE | Score | parameters | GPU time |
| FNN | 15.79 | 148.32 | 119,601 | 165.74 | 9.69 | 77.01 | 2,957,601 | 410.36 |
| SRNN | 15.78 | 541.95 | 82,351 | 117.32 | 12.76 | 99.90 | 3,145,101 | 840.90 |
| GRU | 12.28 | 159.06 | 120,851 | 313.88 | 9.71 | 139.22 | 3,791,351 | 3250.49 |
| LSTM | 15.62 | 188.94 | 139,201 | 319.89 | 11.48 | 103.74 | 4,097,601 | 1672.46 |
| BiLSTM | 13.13 | 532.58 | 338,301 | 261.65 | 11.74 | 222.39 | 9,195,101 | 4673.24 |
| CNN | 19.94 | 969.98 | 45,971 | 22.64 | 12.49 | 287.21 | 2,957,601 | 410.36 |
| CNLSTM | 17.27 | 803.29 | 147,833 | 291.37 | 12.55 | 240.95 | 3,616,601 | 2012.33 |
| SAN | 21.31 | 199.09 | 38,112 | 412.11 | 30.59 | 2735.31 | 56,751 | 853.65 |

The experiment presented here clearly shows that all the multi-head deep learning models evaluated have significant improvement in the RUL prediction compared to the single head implementation. However, the results show that the multi-head architecture also imposes a significantly higher computational burden than the single head design.

### 4.2. The effect of attention on multi-head models

This subsection discusses the comparison result of the stand-alone multi-head model and attention-based multi-head model. This is to demonstrate the performance improvement presented by the attention mechanism. First, considering the different types of attention mechanisms presented in Section 2, two kinds of attention mechanisms are evaluated in this section: soft-multiplicative attention (soft*) and hard multiplicative (hard*) attention. The evaluated attention mechanisms are selected based on their multi-head reproducibility using the Keras on TensorFlow framework.

It is seen in Tables 4 and 5 that there is no clear distinction in the performance of the multi-head soft* attention models and that of the hard* attention model. Also, the empirical values do not show any consistency in the performance of the models. For instance, in Table 4, the SRNN, GRU, SAN multi-head model with soft* attention has 2.95%, 14.59%, and 2.25% reduction in prediction error (RMSE), but have 16.87%, 48.86%, and 12.479% increase in the score value, which shows inconsistency in performance across the evaluation metrics. The FNN and CNN multi-head models with hard* attention show better RUL prediction, with a 5.81% and 16.15% reduction in RMSE, 15.023%, and 28.66% reduction in score value. In contrast, examining the multi-head LSTM, the hard* architecture has a 2.46% reduction in RMSE and 27.71% increase in score value. For BiLSTM and CNLSTM, the hard* attention has a 1.52%, and 2.0% reduction in error, but 6.92% and 14.035 increase in score, respectively. The analysis shows that the soft* LSTM, BiLSTM, and CNLSTM models have a better score value, but worse RMSE. The reverse is the case of multi-head FNN and CNN models with hard* attention having a better RMSE and score.

Table 4: Performance evaluation of multi-head models on FD001 with different types of attention

| Model | Soft* | | | Hard* | | |
| --- | --- | --- | --- | --- | --- | --- |
| | RMSE | Score | GPU | RMSE | Score | GPU |
| FNN | 11.35 | 48.58 | 769.04 | 10.69 | 41.28 | 1026.55 |
| SRNN | 11.54 | 47.64 | 1209.37 | 11.88 | 55.68 | 916.35 |
| GRU | 11.10 | 35.51 | 915.40 | 12.72 | 52.86 | 1718.69 |
| LSTM | 12.19 | 42.33 | 1045.55 | 11.89 | 54.06 | 815.21 |
| BiLSTM | 10.52 | 38.42 | 2493.39 | 10.68 | 35.76 | 1841.13 |
| CNN | 13.81 | 67.30 | 912.62 | 11.58 | 48.01 | 925.71 |
| CNLSTM | 12.49 | 44.03 | 1901.58 | 12.24 | 50.21 | 1407.83 |
| SAN | 26.17 | 778.61 | 1108.98 | 26.76 | 875.78 | 636.21 |

Table 5: Performance evaluation of multi-head models on FD003 with different types of attention

| Model | Soft* | | | Hard* | | |
| --- | --- | --- | --- | --- | --- | --- |
| | RMSE | Score | GPU | RMSE | Score | GPU |
| FNN | 15.60 | 344.80 | 1505.68 | 15.69 | 400.36 | 1733.77 |
| SRNN | 12.29 | 92.84 | 1810.34 | 11.65 | 73.95 | 1885.03 |
| GRU | 12.50 | 248.47 | 1753.18 | 11.02 | 48.23 | 1901.71 |
| LSTM | 10.06 | 69.31 | 1794.31 | 12.10 | 172.96 | 1554.48 |
| BiLSTM | 11.12 | 92.52 | 2508.46 | 9.32 | 56.82 | 4398.29 |
| CNN | 12.18 | 76.86 | 1861.98 | 11.39 | 67.16 | 1309.28 |
| CNLSTM | 12.04 | 121.40 | 1931.54 | 13.43 | 199.85 | 2354.11 |
| SAAN | 30.59 | 2735.31 | 853.65 | 32.39 | 2926.44 | 593.42 |

A similar trend is observed in the evaluation result using the FD003 dataset. For a better perspective of the effect of the attention mechanism, further analysis is done to compare the multi-head model with and without attention. For proper analysis, the soft* attention multi-head model is compared with the multi-head deep learning models without attention across the two datasets, as shown in Tables 6 and 7.

Table 6: Performance evaluation of different multi-head models on FD001 with (W/A) and without (W/A) soft-multiplicative attention

| Model | RMSE | | Score | | Parameters | | GPU time | |
|---|---|---|---|---|---|---|---|---|
| | W/O | W/A | W/O | W/A | W/O | W/A | W/O | W/A |
| FNN | 8.68 | 11.35 | 28.43 | 48.58 | 2,011,201 | 2,053,718 | 290.04 | 769.04 |
| SRNN | 9.99 | 11.54 | 35.31 | 47.64 | 2,138,701 | 2,181,218 | 739.49 | 1209.37 |
| GRU | 10.06 | 11.10 | 34.61 | 35.51 | 2,578,151 | 2,620,668 | 2044.05 | 915.40 |
| LSTM | 9.93 | 12.19 | 29.93 | 42.33 | 2,786,401 | 2,828,918 | 1281.66 | 1045.55 |
| BiLSTM | 11.15 | 10.52 | 39.14 | 38.42 | 6,252,701 | 6,422,718 | 2472.02 | 2493.39 |
| CNN | 8.74 | 13.81 | 26.60 | 67.30 | 2,745,125 | 2,814,774 | 390.84 | 912.62 |
| CNN+LSTM | 11.62 | 12.49 | 54.22 | 44.03 | 2,459,321 | 2,501,838 | 1371.07 | 1901.58 |

Table 7: Performance evaluation of different multi-head models on FD003 with (W/A) and without (W/O) soft-multiplicative attention

| Model | RMSE | | Score | | Parameters | | GPU time | |
|---|---|---|---|---|---|---|---|---|
| | W/O | W/A | W/O | W/A | W/O | W/A | W/O | W/A |
| FNN | 9.69 | 15.60 | 77.01 | 344.80 | 2,957,601 | 3,020,126 | 410.36 | 1505.68 |
| SRNN | 12.76 | 12.29 | 99.90 | 92.84 | 3,145,101 | 3,207,626 | 840.90 | 1810.34 |
| GRU | 9.71 | 12.50 | 159.06 | 248.47 | 3,791,351 | 3,853,876 | 3250.49 | 1753.18 |
| LSTM | 11.48 | 10.06 | 103.74 | 69.31 | 4,097,601 | 4,160,126 | 1672.46 | 1794.31 |
| BiLSTM | 11.74 | 11.12 | 532.58 | 92.52 | 9,195,101 | 9,445,126 | 4673.24 | 2508.47 |
| CNN | 12.49 | 12.18 | 287.21 | 76.86 | 4,036,901 | 4,139,326 | 562.09 | 1861.98 |
| CNN+LSTM | 12.55 | 12.04 | 240.96 | 121.40 | 3,616,601 | 3,679,126 | 2012.33 | 1931.54 |

Tables 6 and 7 show that, for most of the multi-head deep learning models evaluated, the attention mechanism performs worse than the multi-head model without attention. As seen in Table 6, on FD001 dataset, FNN, SRNN, GRU, LSTM, and CNN without attention have 30.76%, 15.52%, 10.34%, 22.76%, 58.01% reduction in RMSE, and 70.87%, 34.92%, 2.60%, 41.43%, 153% reduction in the score value respectively, while CNLSTM without attention have 7.49% reduction in RMSE but 18.79% increase in score value. The BiLSTM with attention has 5.65% reduction in RMSE and 1.84% reduction in score value, making the only multi-head-attention model with consistent improvement.

For dataset FD003 result in Table 7, the FNN and GRU multi-head models without attention have 60.99% and 28.73% reduction in the RMSE, and 347.73%, and 56.21% reduction in the score value respectively, while multi-head SRNN, LSTM, BiLSTM, CNN, and CNLSTM models with attention have 3.68%, 12.37%, 5.28%, 2.48%, and 4.06% reduction in RMSE, while SRNN and

LSTM with attention have 7.07%, 33.19% reduction in the score values, and BiLSTM, CNN, and CNNLSTM have 82.62%, 73.24%, and 49.62% increase in the score value between the model with and without attention respectively.

This experimental result shows that the attention mechanism does not necessarily improve RUL predictive performance. The results show that for most multi-head deep learning architecture, models without attention is sufficient to capture the information inherent in the time series dataset utilized for RUL prediction. The result also shows that a stack of multi-head fully connected neural networks ( has the best predictive performance for the FD001 and FD003 datasets. To support this conclusion, the section below shows the comparison of the state-of-the-art single head models with the best multi-head architecture presented in this work.

### 4.3. Comparison of multi-head models with the state-of-the-art

Table 8 compares the best multi-head model result with other state-of-the-art approaches on the two turbofan engine benchmark datasets. In table 8, the state-of-the-art models compared are the deep convolution neural network (DCNN (Li, Ding et al. 2018)), generative adversarial network (DCGAN (Hou, Xu et al. 2020)), restricted Boltzmann machine with long short-term memory (RBM+LSTM (Ellefsen, Bjørlykhaug et al. 2019)), bidirectional LSTM (BiLSTM (Yu, Kim et al. 2019)), and causal augmented temporary convolution network (CaConvNet (Ayodeji, Wang et al. 2021)).

Table 8: Performance evaluation of the multi-head FNN model with state-of-the-art deep learning models

| Model | Source | FD001 | | FD003 | |
|---|---|---|---|---|---|
| | | Score | RMSE | Score | RMSE |
| DCNN | RESS | 273.70 | 12.61 | 284.10 | 12.64 |
| DCGAN | CIN | 174 | 10.71 | 273 | *11.48* |
| RBM+LSTM | RESS | 231 | 12.56 | 251 | 12.10 |
| BiLSTM+ED | MSSP | 273 | 14.47 | 574 | 17.48 |
| CaConvNet | ISA T. | 84.83 | 11.83 | **55.52** | **9.24** |
| Multi-head FNN | **Current work** | **28.43** | **8.68** | 77.01 | 9.69 |

It is seen in Table 8 that the multi-head model, FNN, performs better than the current best state of the art model, on dataset FD001, with 66.49% improvement in RMSE and 26.63% improvement in the score value. On the FD003 dataset, the multi-head FNN also performs better than most state-of-the-art models, except the CaConvNet model with a 38.71% reduction in the RMSE value and a 4.87% reduction in the score value. Although the FNN model has more computation burden (i.e. compared with CaConvNet with 466,333 parameters), the RUL prediction result from the multi-head FNN clearly shows performance improvement. To further demonstrate the improved performance of the multi-head model, Fig 13(a)-(d) shows the plot of predicted RUL vs. true RUL for selected multi-head models.

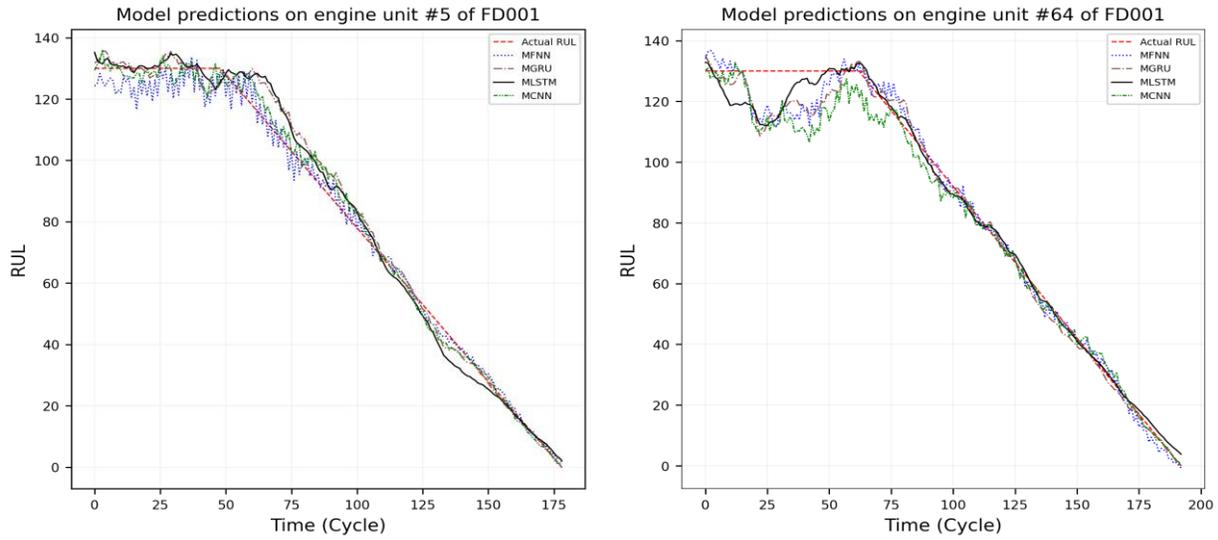

(a)             (b)

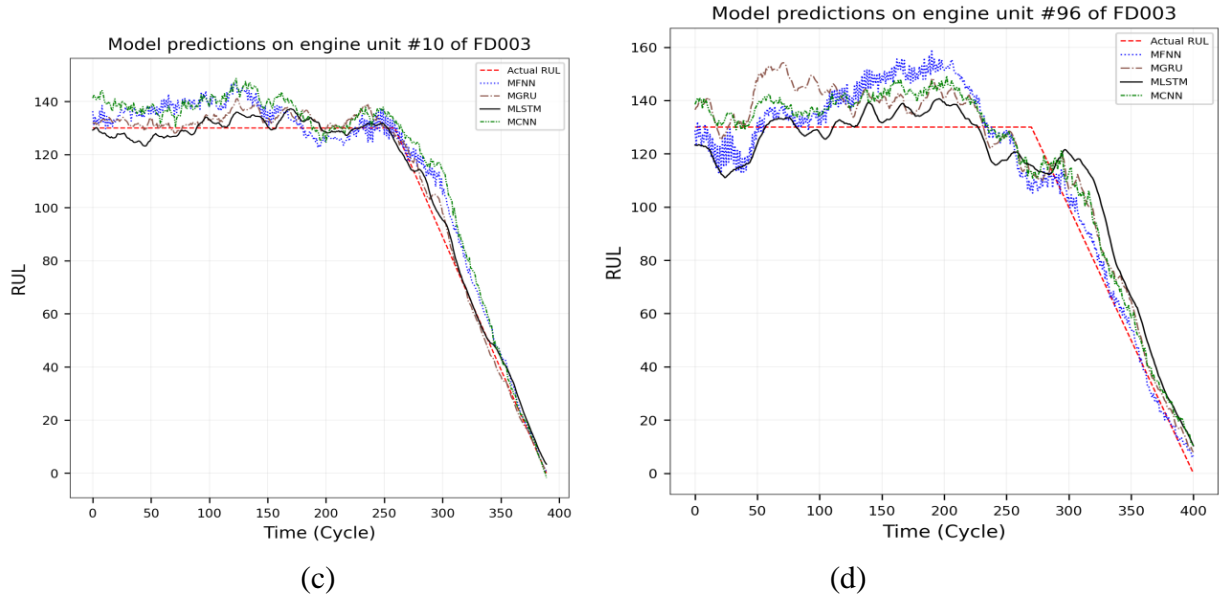

(c)                      (d)

Fig.13: (a)&(b) Performance of selected multi-head models on engine #5 and #64 of FD001 dataset; (c) & (d) on engine #10 and #36 of FD003 dataset

## 5. Conclusion

No comparative study critically explores the gap between the emerging architectures such as multi-head and attention mechanisms and the well-established conventional single head approach. This work discusses an extensive experiment on multi-head attention-based deep learning models, and analyze the benefit of using a context-specific head to independently capture the inherent pattern in each signal in multivariate time series. The models are evaluated on two subsets in the C-MAPSS dataset.

The evaluation results show that the multi-head model developed with attention-based multilayer perceptron performs best on both subsets. It is also observed that adding attention layers does not improve the prognostic performance of some models. The experimental results allow the conclusion that the use of multi-head models gives consistently better results for RUL prediction. The result also shows that utilizing the attention mechanism is task-sensitive and model-dependent, as it does not provide consistent improvement across the dataset used. The empirical evaluation results show the importance of multi-head models for critical system safety service life estimation, end of life prediction, and continuity in the industrial process by enhancing part replacement strategy. The result is also valuable for understanding and improving the remaining useful life of turbofan aircraft engines using multivariate degradation information. Nevertheless, it is worth mentioning that the FD001 and FD003 datasets used in this experiment may not represent the complexity in most industrial assets. Hence more investigation is required to properly evaluate the effect of attention mechanism to capture the information in datasets that defines complex scenarios or reflect the complexity in industrial assets.

**CRediT author statement**
**Abiodun Ayodeji:** Conceptualization, Methodology, Data curation, Writing- Original draft, Software. **Wenhai Wang**: Visualization, Investigation. **Jianzhong Su:** Supervision. **Jianquan Yuan**: Supervision, Validation, **Xinggao Liu**: Supervision, Investigation, Writing-review, and editing.

**Declaration of competing interest**

The authors declare that they have no known competing financial interests or personal relationships that could have appeared to influence the work reported in this paper

**Acknowledgment**
This work is supported by the National Natural Science Foundation of China (62073288, 12075212), National Key R&D Program of China (Grant No. 2018YFB2004200) and the Fundamental Research Funds for the Central Universities (Zhejiang University NGICS Platform) and their supports are thereby acknowledged.